\documentclass[journal, twoside]{IEEEtran}
\IEEEoverridecommandlockouts
\usepackage{cite}
\usepackage{hyperref}
\usepackage{amsmath,amssymb,amsfonts}
\usepackage{algorithmic}
\usepackage{graphicx}
\usepackage{textcomp}
\usepackage{xcolor}
\def\BibTeX{{\rm B\kern-.05em{\sc i\kern-.025em b}\kern-.08em
    T\kern-.1667em\lower.7ex\hbox{E}\kern-.125emX}}
\begin{document}

\title{The Hardware Impact of Quantization and Pruning for Weights in Spiking Neural Networks}

\author{
 %
 \IEEEauthorblockN{ Clemens JS Schaefer, Pooria Taheri, Mark Horeni, and Siddharth Joshi}
 
 \IEEEauthorblockA{Department of Computer Science and Engineering, University of Notre Dame, Notre Dame, IN, USA } 
 
Email: {\{cschaef6, ptaheri, mhoreni, sjoshi2\}@nd.edu}
\thanks{This work was supported in part by ASCENT, one of six centers in JUMP, a SRC program sponsored by DARPA.}
}

\makeatletter
\def\ps@IEEEtitlepagestyle{
  \def\@oddfoot{\mycopyrightnotice}
  \def\@evenfoot{}
}
\def\mycopyrightnotice{
  {\footnotesize
  \begin{minipage}{\textwidth}
  \copyright~2023 IEEE.  Personal use of this material is permitted.  Permission from IEEE must be obtained for all other uses, in any current or future media, including reprinting/republishing this material for advertising or promotional purposes, creating new collective works, for resale or redistribution to servers or lists, or reuse of any copyrighted component of this work in other works.
  \end{minipage}
  }
}
\maketitle
\IEEEpeerreviewmaketitle

\begin{abstract}
Energy efficient implementations and deployments of Spiking neural networks (SNNs) have been of great interest due to the possibility of developing artificial systems that can achieve the computational powers and energy efficiency of the biological brain. Efficient implementations of SNNs on modern digital hardware are also inspired by advances in machine learning and deep neural networks (DNNs). Two techniques widely employed in the efficient deployment of DNNs -- the quantization and pruning of parameters, can both compress the model size, reduce memory footprints, and facilitate low-latency execution. The interaction between quantization and pruning and how they might impact model performance on SNN accelerators is currently unknown. We study various combinations of pruning and quantization in isolation, cumulatively, and simultaneously (jointly) to a state-of-the-art SNN targeting gesture recognition for dynamic vision sensor cameras (DVS). We show that this state-of-the-art model is amenable to aggressive parameter quantization, not suffering from any loss in accuracy down to ternary weights. However,  pruning only maintains iso-accuracy up to 80\% sparsity, which results in 45\% more energy than the best quantization on our architectural model. Applying both pruning and quantization can result in an accuracy loss to offer a favourable trade-off on the energy-accuracy Pareto-frontier for the given hardware configuration.\footnote{Code \url{https://github.com/Intelligent-Microsystems-Lab/SNNQuantPrune}}
\end{abstract}


\section{Introduction}

Spiking neural networks (SNNs) offer a promising way of delivering low-latency, energy-efficient, yet highly accurate machine intelligence in resource-constraint environments. By leveraging bio-inspired temporal spike communication between neurons, SNNs promise to  reduce communication and compute overheads drastically compared to their rate-coded artificial neural network (ANN) counterparts. In particular, SNNs are attractive for temporal tasks such as event-driven audio and vision classification problems~\cite{amir2017low,cramer2020heidelberg}. Because spike storage and communication are often low-overhead (compared to DNN activations), the primary energy expenditure for SNN accelerators can be attributed to the cost of accessing and computing with high-precision model parameters.

Parameter quantization and parameter pruning are widely used to reduce model footprint~\cite{dong2020hawq, zhuang2018discrimination}. Quantization, especially fixed-point integer quantization, can enable the use of cheaper lower precision arithmetic hardware and simultaneously reduce the required memory bandwidth for accessing all the parameters in a layer. On the other hand, pruning compresses a model by identifying unimportant weights and zeroing them out. Judicious use of both is critical in practical deployments of machine intelligence when the system might be constrained by size, weight, area and power (SWAP). Developing a better understanding of the interaction between quantization and pruning is critical to developing the next generation of SNN accelerators.

Efforts to quantize SNN parameters have recently gained traction. Authors in~\cite{lui2021hessian} developed a second-order method to allocate bit-widths per-layer an reduce the model size by 58\% while only incurring an accuracy loss of 0.3\% on neuromorphic MNIST. Other work~\cite{putra2021q} demonstrated up to a $2\times$ model size reduction using quantization with minimal (2\%) loss in accuracy on the DVS gesture data set. Separately, the authors in ~\cite{schaefer2020quantizing} demonstrated quantization in both inference and training, allowing for a memory footprint reduction of 73.78\% with approx. 1\% loss in accuracy on the DVS gesture data set. Similarly, authors in~\cite{eshraghian2022fine} showed that SNNs with binarized weights can still achieve high accuracy at tasks where temporal features are not task-critical (e.g. DVS Gesture $0.69\%$ accuracy loss). They note that while on tasks highly dependent on temporal features such binarization incurs a greater degradation in accuracies hits (e.g. Heidelberg digits $16.16\%$ accuracy loss). Authors in~\cite{sorbaro2020optimizing} directly optimize the energy efficiency of SNNs obtained by conversion from ANNs by adding an activation penalty and quantization-aware training to the ANN training. They demonstrated a 10-fold drop in synaptic operations while only facing a minimal (1-2\%) loss in accuracy on the DVS-MNIST task.

The authors in~\cite{shi2019soft} demonstrated a 75\% reduction in weights through pruning while achieving a~90\% accuracy on the MNIST data set. More recently, researchers combined weight and connectivity training of SNNs~\cite{chen2021pruning} to reduce the number of parameters in a spiking CNN to $\approx1\%$ of the original while incurring an accuracy loss of $~3.5\%$  on the CIFAR-10 data set.

There has been limited work on the interaction between pruning and quantization with recent work in~\cite{chowdhury2021spatio} showing $5$-bit parameter quantization and $10-14\times$ model compression through pruning. However, most work in this domain has not focused on temporal tasks, where spike times and neuron states are critical to SNN functionality. This paper focuses comprehensively on the energy efficiency derived from quantizing and pruning a state-of-the-art SNN~\cite{zhu2022tcja}.

\section{Methods}

\subsection{Spiking Neural Networks}
Spiking neural networks are motivated by biological neural networks, where both communication and computation use action potentials. Similar to the work in~\cite{zhu2022tcja}, we use a standard leaky-integrate and fire neuron, written as:

\begin{align}
    u(t)^l_i &= \tau  u(t-1)^l_i + \sum_j s(t)^{l-1}_{ij} w^l_{ij}, \nonumber \\ 
    s(t)^l_i &= H(u(t)^l_i - V_{th}).\label{eq:snn}
\end{align}

Where $u(t)^l_i$ is the membrane potential of the $i^{th}$ neuron in layer $l$ at time step $t$ and decays by the factor $\tau$ while accumulating spiking inputs $s(t)^{l-1}_{ij}$ from the previous layer weighted by $w_{ij}$. We do not constrain the connectivity of these layers. When the membrane potential exceeds the spike threshold $V_{th}$, the neuron issues a spike (see  eq.~\eqref{eq:snn}) through the Heaviside step function $H$ and resets its membrane potential to a resting potential (set to $0$ in this work). 

\begin{figure}[htb]
\centering
  \includegraphics[width=.85\columnwidth]{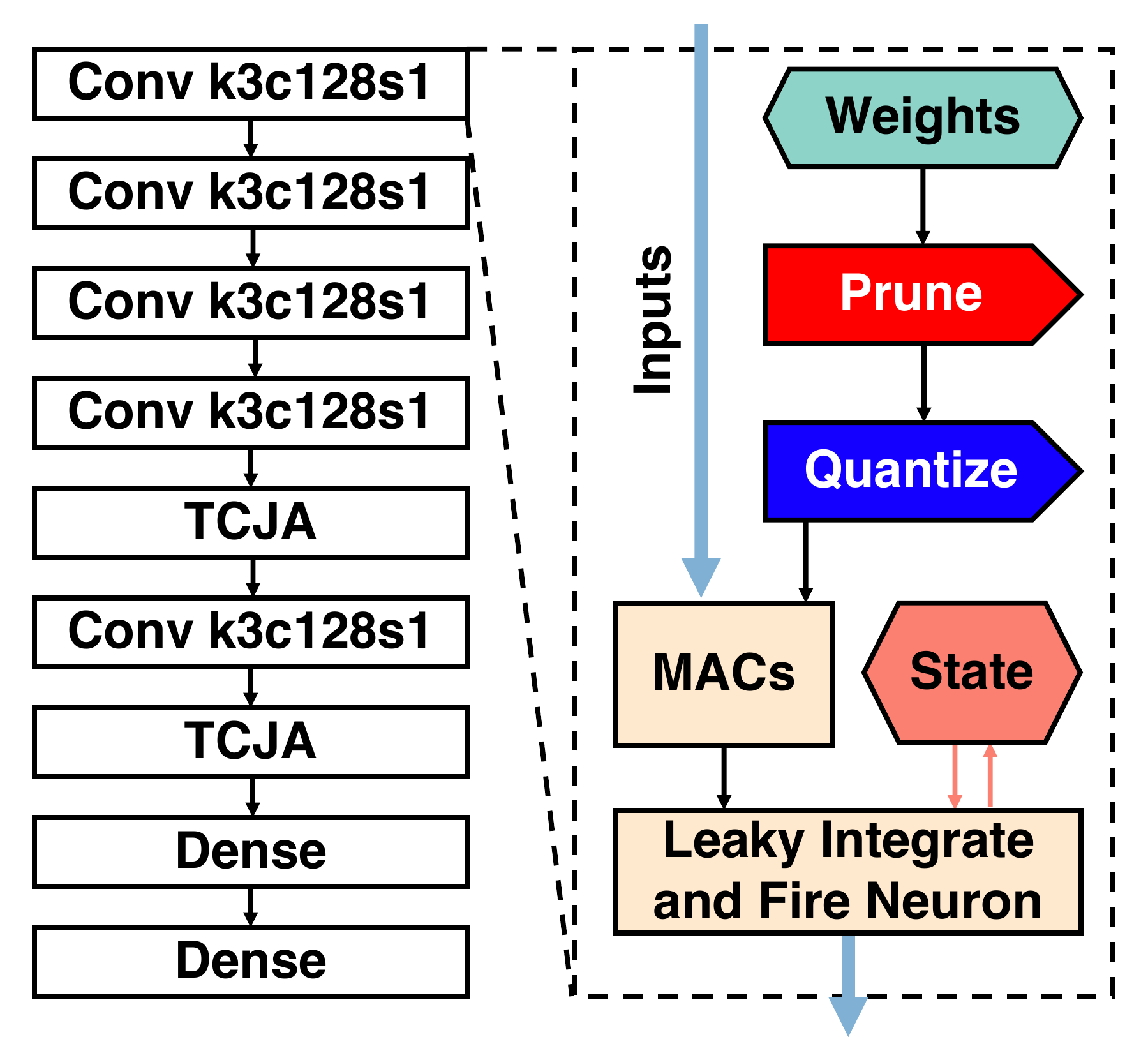}
\caption{On the left the schematic SNN architecture proposed by~\cite{zhu2022tcja} using stacked spiking convolution blocks with $3\times3$ weight kernels, 128 output channels and striding of 1. Followed by temporal-channel joint attention (TCJA) blocks and spiking dense blocks for classification. On the right is a simplified flow within a spiking layer: weights are first pruned and then quantized (preprocessing), inputs are multiplied and accumulated with preprocessed weights before being fed into the logic of the leaky integrate and fire neurons, which additionally relies on a stored state (membrane potential).}
\label{fig:arch_nn}
\end{figure}
We illustrate our exemplar SNN model in~\ref{fig:arch_nn}, with  max pooling and batch norm layers (in the convolutional layers) omitted for clarity. The model is trained using backpropagation-through-time with arc-tangent surrogate gradients~\cite{neftci2019surrogate, lee2016training}.
\subsection{Quantization}
In this work, we only quantify the effects of quantizing synaptic weights, employing standard quantization techniques like clipping and rounding as described in~\cite{park2020profit}: 
\begin{equation} \label{eq:quant}
    Q_u(x) = \mathrm{clip}\left( \mathrm{round}\left( \frac{x}{s_{\text{in}}} \right) , -2^{b-1}, 2^{b-1}-1 \right)  \cdot s_{\text{out}}.\nonumber
\end{equation}
Here, $Q_u$ is the quantizer function, $b$ is the number of bits allocated to the quantizer, and $x$ is the value to be quantized. Additionally, $s_{\text{in}}$ and $s_{\text{out}}$ are learnable scale parameters. The scale parameters are initially set to $3\times$ the standard deviation of the underlying weight distribution added to its mean. In this work, we implement quantization-aware training through gradient scaling~\cite{lee2021network} to minimize accuracy degradation.

\subsection{Pruning}
We implement global unstructured pruning by enforcing a model-wide sparsity target ($\omega$) across all the SNN model parameters (weights). During pruning, we rank the weights by magnitude and prune the $\omega\%$ lowest.

\subsection{Energy Evaluation}

\begin{figure}[htb]
\centering
  \includegraphics[width=.65\columnwidth,trim={0 0 0 0},clip]{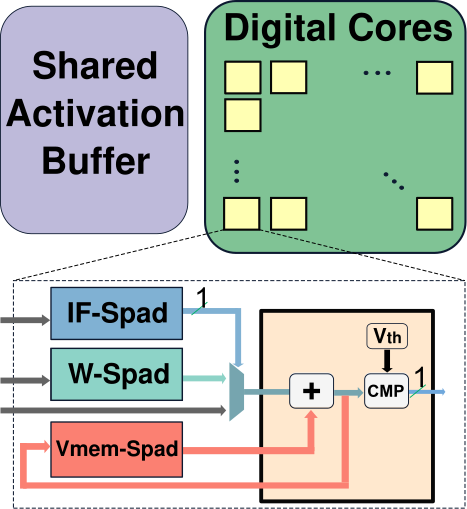}
\caption{Eyeriss-like spiking architecture consisting of a shared activation buffer connected to an array of digital cores. A single digital core has dedicated memory for inputs (IF-Spad), weights (W-Spad), state (Vmem-Spad) and spiking threshold ($V_{th}$). The digital cores possess arithmetic components, e.g. an accumulator or a comparator (CMP) for spike generation.}
\label{fig:arch_eye}
\end{figure}

We use an Eyeriss-Like architecture~\cite{isscc_2016_chen_eyeriss} illustrated in Figure~\ref{fig:arch_eye}. This accelerator implements a grid of digital processing elements (PEs) with input spikes and membrane partial sums supplied from a shared activation buffer. Each digital core has scratchpad memories for input spikes, weights, neuron state, and spiking threshold. The PEs can leverage multiple sparse representation schemes, incorporating features like clock gating and input read skipping. These sparsity-aware techniques are applied for input spikes, weights, and output spikes. Across the PE grid, inputs can be shared diagonally, weights can be multicasted horizontally, and partial sums can be accumulated vertically. We removed multipliers from the PE and modified the input scratch pads to be only 1-bit wide. This architecture is evaluated for our SNN workload of choice using high-level component-based energy estimation tools~\cite{iccad_2019_accelergy, parashar2019timeloop} calibrated to a commercially available 28~nm CMOS node. We determine spike activity statistics through the training-time sparsity. Tensor-sparsity statistics and calibrated energy costs are used for loop analysis, where loop interchange analysis is employed to search for efficient dataflows~\cite{parashar2019timeloop}. All results enforce causality, preventing temporal loop reordering.

\section{Results}
\begin{figure}[htb]
  \includegraphics[width=.9\linewidth]{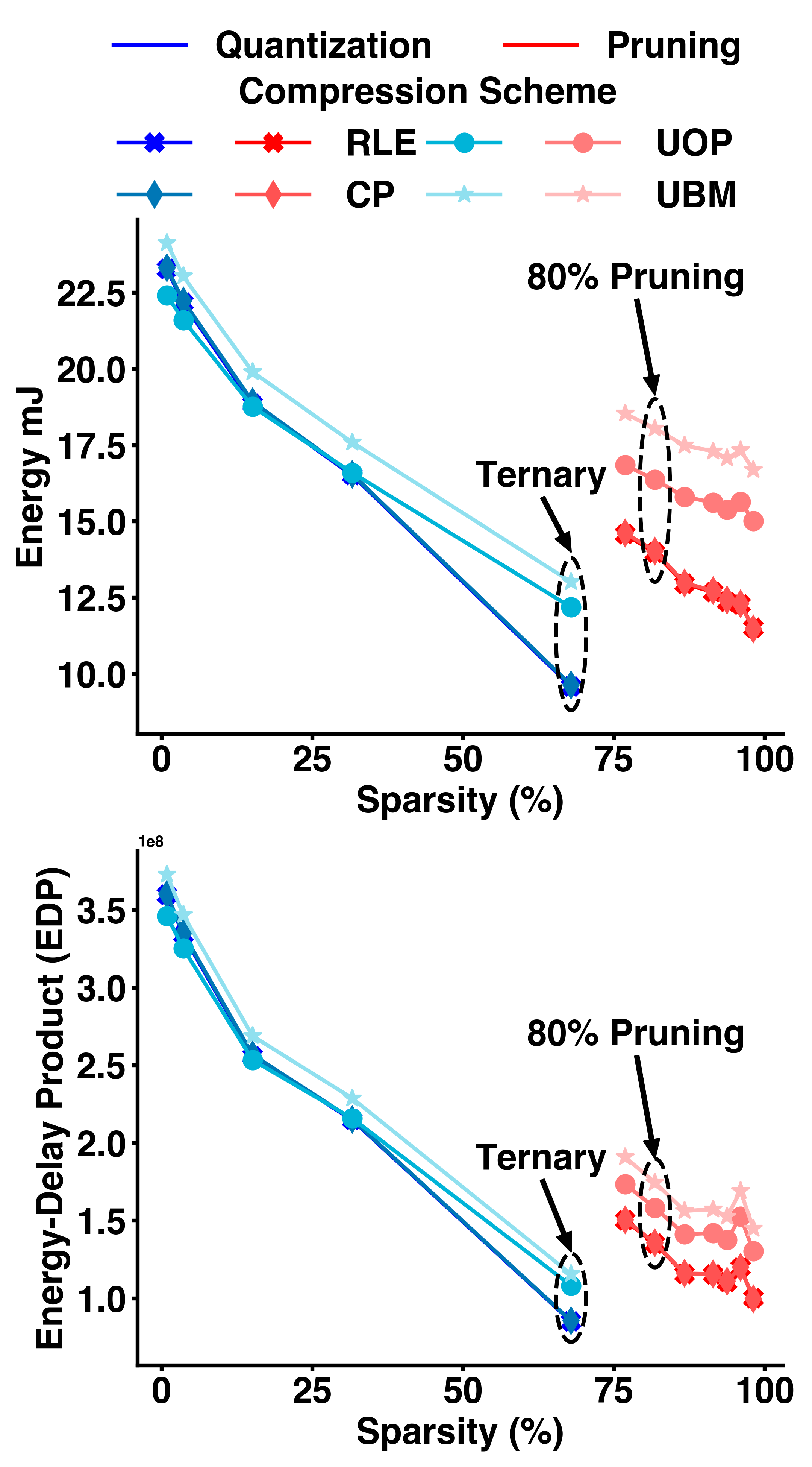}
\caption{Effect of sparsity on energy (top) and energy-delay product (EDP) (bottom). We compute the sparsity (compared to a fully dense model) of our SNN model after quantization-aware and pruning-aware fine-tuning. For pruning-aware training, we ensure that although the weight updates can set more weights to zero the pruning mask prevents the training from regrowing connections. For this guided pruning, we set the sparsity targets to 75\%, 80\%, 85\%, 90\%, 92.5\%, 95\% and 97.5\%. We evaluated weight quantization across 8b, 6b, 4b, 3b, and ternary precision. We measure quantization-induced weight sparsity for these models by detecting how many values are set to 0 after quantization and finetuning. We explore different compression schemes: uncompressed bitmask (UBM), uncompressed offset pair (UOP), coordinate payload (CP), and run length encoding (RLE). For models with little sparsity (e.g. less than 25\%) the UOP scheme performs best. However, models with considerable amounts of sparsity achieve better energy and EDP performance with RLE. Note although RLE and CP nearly overlap each other, RLE is strictly better, although by a small margin. The highlighted models are ternarized models and pruned models with 80\% sparsity target, all delivering iso-accuracy, demonstrating a clear advantage for ternary quantization.}
\label{fig:sparse_ener}
\end{figure}

We evaluate the pruning and quantization performance of the SNN model~\cite{zhu2022tcja} on the DVS gesture data set~\cite{amir2017low} using spike-count for classification. Our baseline model has weights quantized to 8 bits while delivering 97.57\% accuracy on DVS gestures.  Figure~\ref{fig:arch_nn} shows the SNN architecture and the location in the computational graph where pruning and quantization operations were applied to the weights. Finetuning experiments started with pre-trained floating point weights and lasted 50 epochs with a linear warm-up and cosine decay learning rate schedule with a peak learning rate of 0.001.

\begin{figure}[!tb]
\includegraphics[width=.9\linewidth, trim={.9cm .5cm .4cm .5cm}, clip]{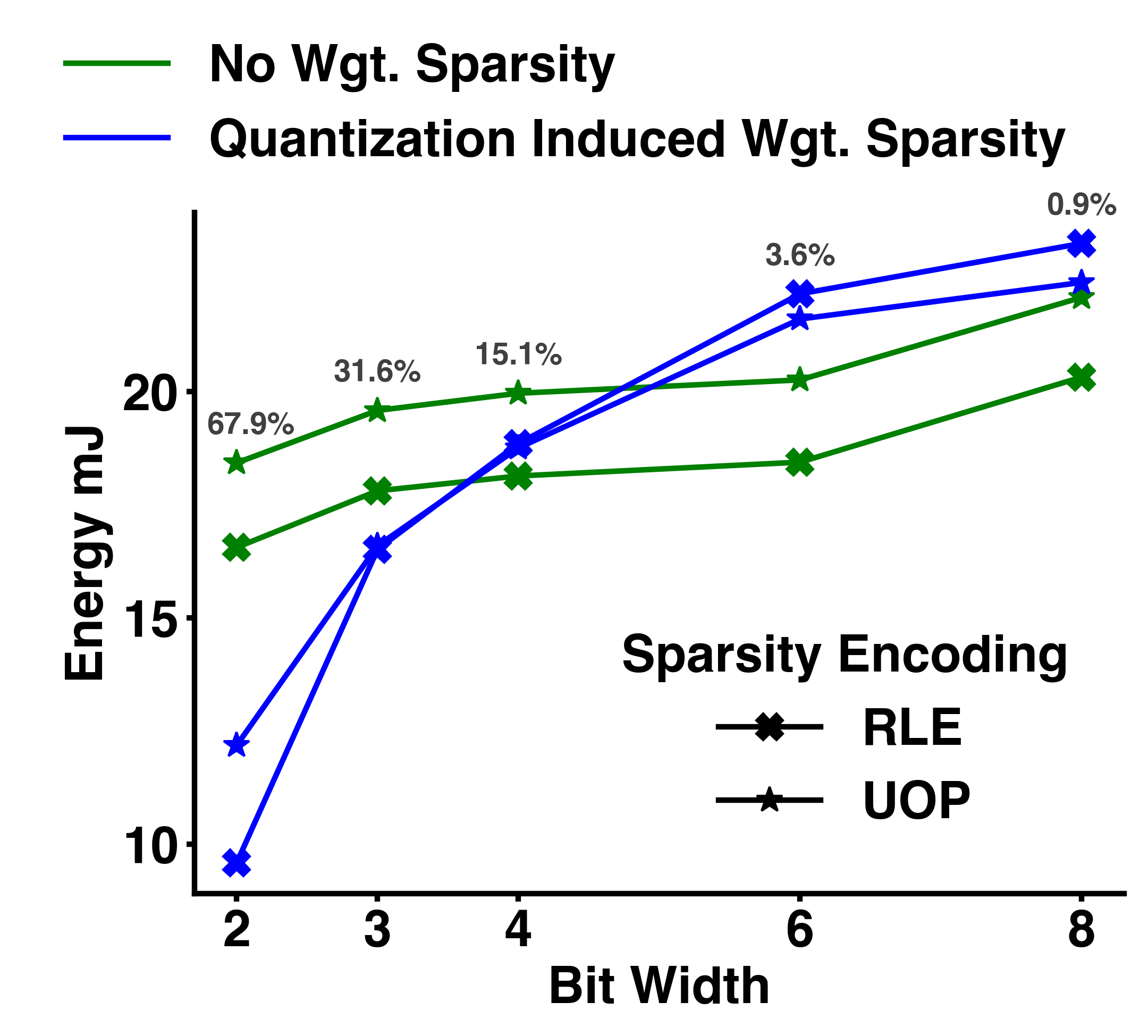}
\caption{Highlighting energy overheads and gains for using sparse representation. Quantization-induced sparsity reduces as model precision increases, leading to applying sparse representations to weights being unsuitable at higher precisions. We still employ sparse representation schemes for input and output spikes to better leverage activity sparsity in SNNs. We also annotate the quantization-induced weight sparsity levels above the data points.}
\label{fig:sp_qu_gain}
\end{figure}

Figure~\ref{fig:sparse_ener} illustrates how different rates of sparsity achieved through pruning and quantization interact with the different compression schemes studied. These include uncompressed bitmask (UBM), uncompressed offset pair (UOP), coordinate payload (CP), and run length encoding (RLE). We see that for highly quantized SNNs, such as those resulting from ternarization the accelerator incurs lower energy and latency in computing the SNN operations when compared to pruned SNNs that deliver the same accuracy. In part,  when terenerizing the weights, there are only three representation levels available, and most weights are quantized to $0$. This allows ternarization to simultaneously benefit from both low-precision computing and high weight sparsity and model compression. At higher levels of precision, there is insufficient quantization-induced sparsity for the network to benefit from sparse representation (see upper left of Figs~\ref{fig:sparse_ener}). At higher sparsity rates, RLE outperforms other representation formats. RLE and CP formats incur similar overheads and deliver similar performance in energy and energy-delay-product (EDP), with RLE being, on average, .3\% better on both metrics. The 8b and 6b models for quantization incur similar energy/energy-delay costs, as seen by their clustering in the upper right corner of both plots in Fig.~\ref{fig:sparse_ener}. In contrast, 4b, 3b, and ternary models benefit from both sparsity and reduced numerical precision, with ternarization gaining the most. 

\begin{figure}[!tb]
\includegraphics[width=.9\linewidth, trim={.4cm .5cm .4cm .4cm}, clip]{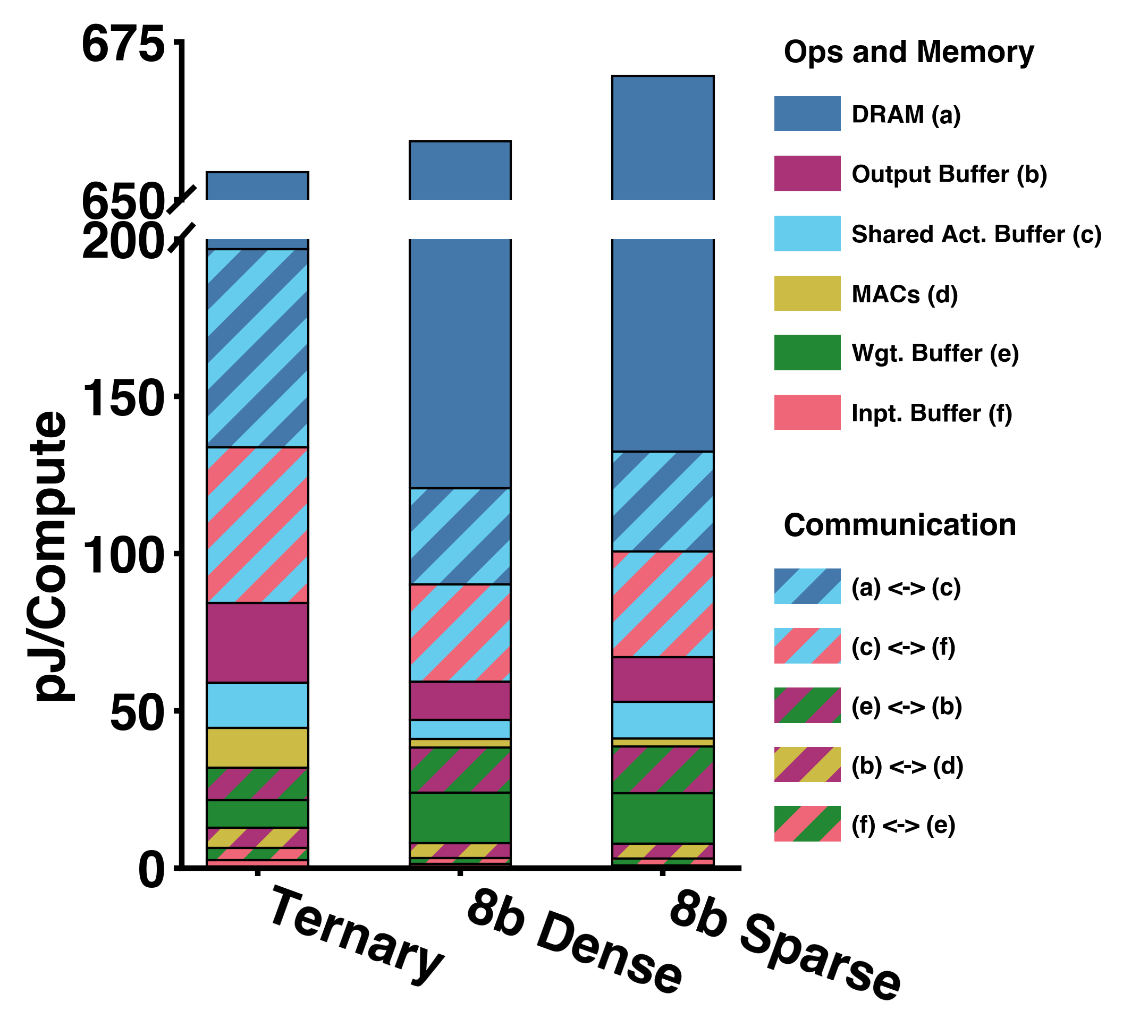}
\caption{A breakdown of normalized energy for different data-movement and operations for three different configurations. First, we see that ternarization incurs minimal DRAM access which is where significant efficiency is gained. Next, for the configuration with 8 bit sparse models, storing and accessing the additional metadata incurs significant costs. This metadata overhead does not exist for the dense 8 bit model. In each case, we employ RLE sparse-storage. Data movement to and from the shared activation buffer incurs the second highest energy costs in our analysis.}
\label{fig:breakdown}
\end{figure}

We better disentangle the benefits of quantization and quantization-induced sparsity, by examining its variation across multiple quantization levels in Fig.~\ref{fig:sp_qu_gain}. For 8 and 6 bit weights the model incurs substantial overheads due to the additional sparse-storage format related metadata. We still employ sparse-storage formats for spike activity, with RLE storage performing better due to the extreme sparsity in spike activity. At the higher precision levels, with a crossover point for 4-b weights, employing RLE and ignoring any sparsity in weights delivers higher performance. However, for 3 bit and weight ternarization, there is a significant improvement to be derived from employing sparse storage formats in weights too. We additionally show the energy breakdown, normalized to computing energy for data movement across the accelerator in Fig.~\ref{fig:breakdown}. The larger energy cost of transferring metadata from DRAM for 8 bit sparse weights results in greater energy incurred for the entire model. Remarkably, the ternarization includes significantly more utilization of the intermediate memories which in turn leads to improved energy-efficiency.

Although both pruning and quantization try to leverage model robustness to improve energy, they operate along different principles. Quantization leverages model overparameterization to facilitate computations at a lower precision while pruning reduces the redundancy in the model parameters to compress it. Consequently, pruning and quantization can often be at odds, requiring a careful study of their interaction. We study two strategies for pruning and quantizing SNNs: i) \textit{cumulatively}, where the model is first finetuned with pruning, and after half the finetuning epochs, we commence quantization aware training, and ii) \textit{jointly}, where the model is simultaneously pruned and quantized (as shown in Fig.~\ref{fig:arch_nn}).

\begin{figure}[htb]
\centering
  \includegraphics[width=.87\linewidth, trim={0 0 0 0}, clip]{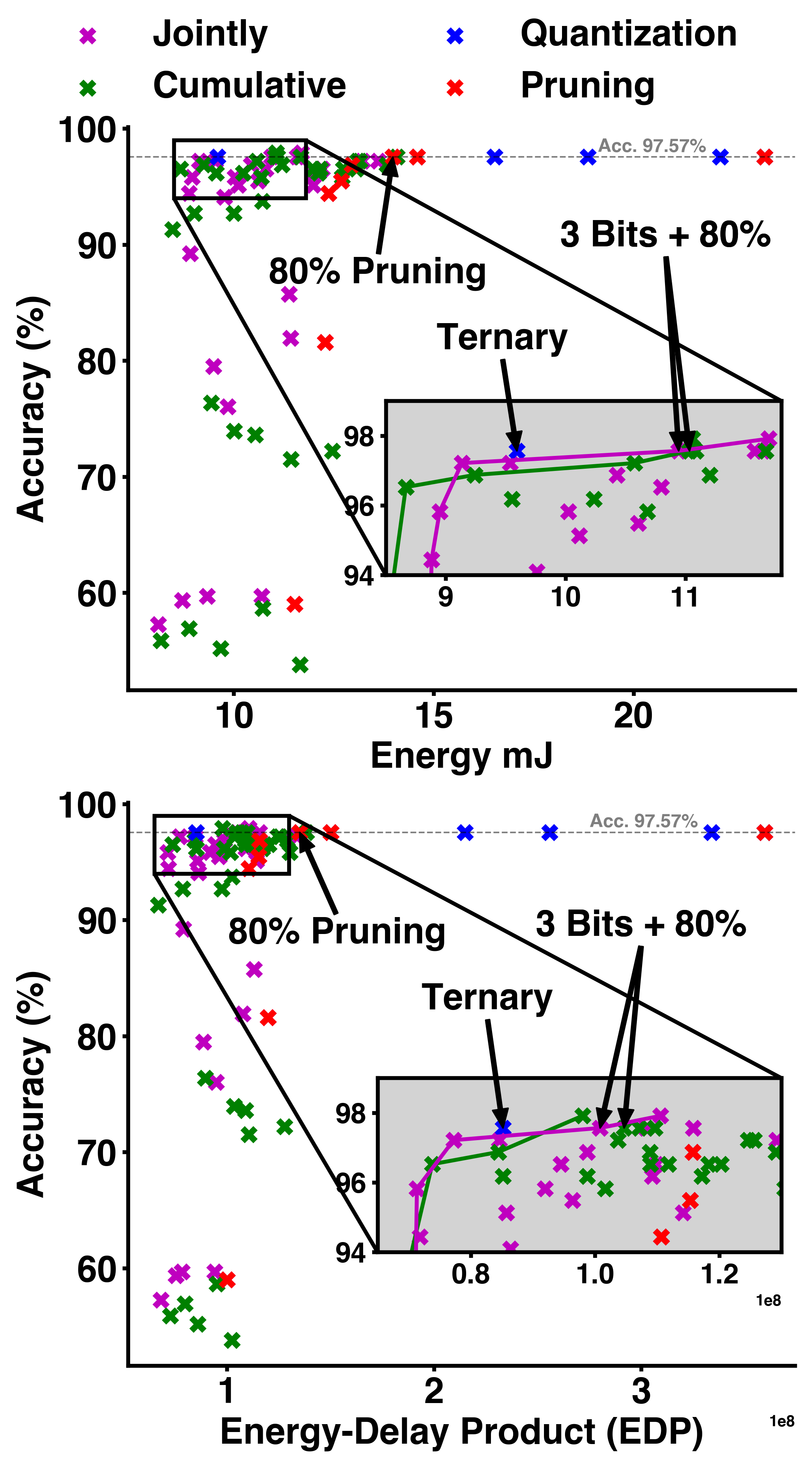}
\caption{Illustration of accuracy vs. energy trade-off (top) and accuracy vs. tradeoff between energy-delay product (bottom) for studied model compression techniques. We study the following interactions between pruning and quantization: quantization only, pruning only, cumulative quantized \& pruned, and jointly quantized \& pruned models. Inset of the figure highlights the points where the models start to degrade from iso-accuracy. Note that all quantized models down to ternary quantization maintain iso-accuracy meanwhile, pruned models degrade at higher energy cost or EDP. Mixed model using both quantization and pruning form the Pareto-curve once model accuracy degrades below iso-accuracy.}
\label{fig:acc_ener}
\end{figure}


Figure~\ref{fig:acc_ener} demonstrates how the different strategies for quantizing and pruning impact the model accuracy and energy-efficiency (top) / energy-delay product (bottom) for our architecture of choice. Although 80\% pruning does not yet suffer from accuracy loss, it is 45\% costlier than ternary quantization, which can maintain accuracy for this model. If constrained to iso-accuracy, a pure quantization approach outperforms all other alternatives, with the 3-bit quantization with 80\% pruning occupying a larger footprint than the ternary network. However, more fine-grained trade-offs between the model accuracy and energy can be achieved across various combined strategies, as shown in the inset of Fig~\ref{fig:acc_ener}. The mixed pruning and quantization schemes enable operation along the Pareto curve when some loss in accuracy can be tolerated, delivering SNN models suitable for multiple SWAP constraints. Cumulative quantization and pruning maintains accuracy in low-energy regimes, however, the overhead of encoding these models obviates this advantage in terms of energy-delay product.

\begin{figure}[!tb]
\includegraphics[width=.9\linewidth, trim={0 1.cm 0 0}]{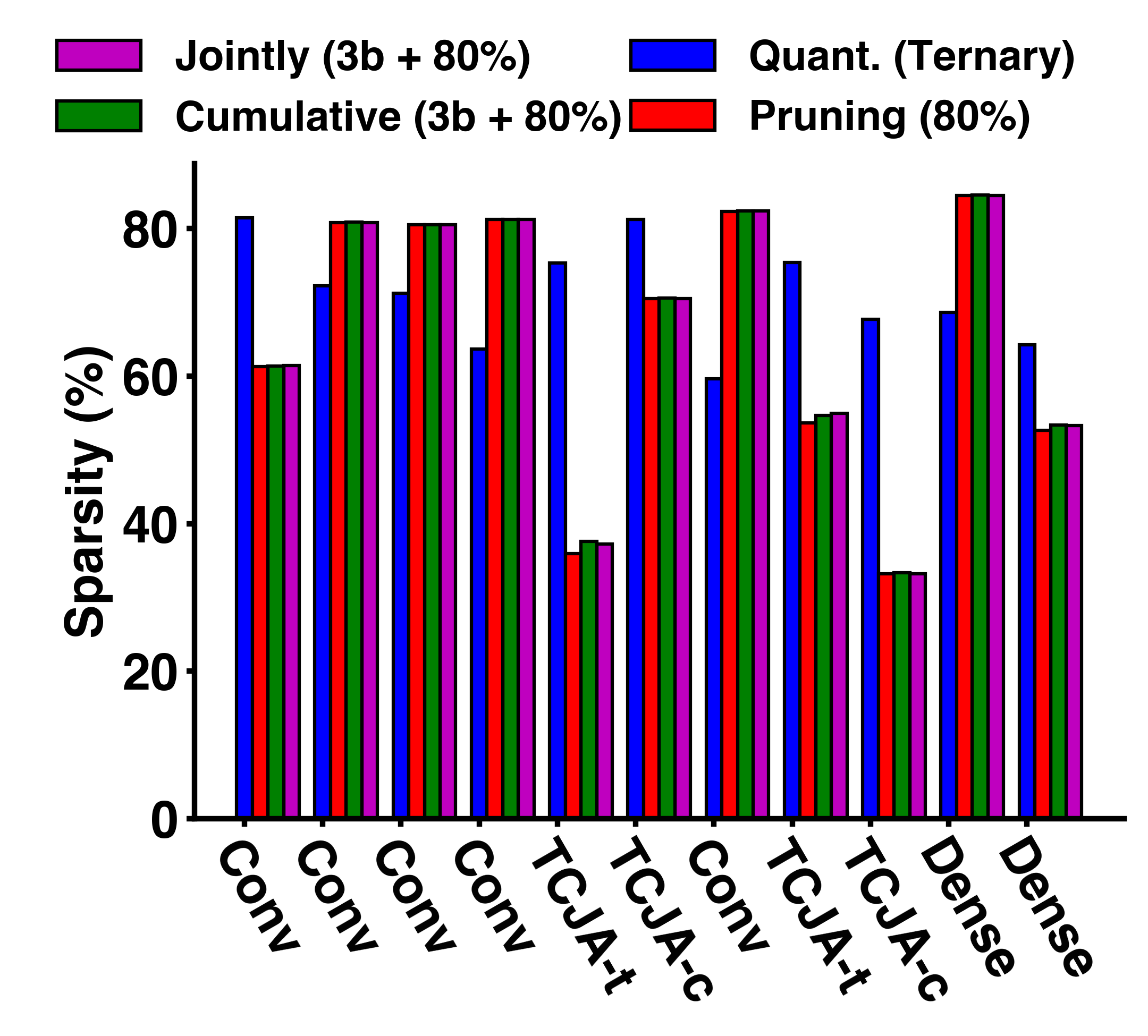}
\caption{Sparsity insight into the SNN model for the smallest iso-accurate models for quantization, pruning, cumulative quantization \& pruning and joint quantization \& pruning. The two mixed approaches replicate the internal sparsity distribution of the pruned model, meanwhile ternary quantization results in a significantly different sparsity distribution (note that the overall sparsity of the ternarized model is 67.9\%).}
\label{fig:sparse}
\end{figure}


We examine how the layer sparsity statistics change across the best iso-accuracy model resulting from quantization, pruning, cumulative quantization \& pruning and joint quantization \& pruning in figure~\ref{fig:sparse}. We observe that the pruning and mixed schemes all have a relatively similar sparsity distribution among layers, with the only difference being that the joint and cumulative approaches can deliver higher sparsity on some layers (e.g., the first TCJA-t). On the other hand, the ternary quantization scheme presents a significantly different sparsity distribution across layers. We propose that the significant energy advantages from terenarization can be attributed to the combination of ternary encoding and consistently higher sparsity levels achieved for the first layer, the last layer, and the two TCJA layers. The different sparsity levels across layers of the iso-accuracy models also hint at commonly observed non-uniform impact for different layer types on energy consumption, e.g. different tensor sizes affecting compiler mappings.

\section{Conclusion}

We analyze the energy-accuracy trade-off between quantization and pruning in state-of-the-art spiking neural networks. We also provide a realistic analysis of how quantization and pruning might interact with a baseline digital SNN accelerator. Our results showed that exploiting quantization-induced sparsity, which is particularly beneficial for weight ternarization, can lead to remarkable performance benefits. By carefully employing such aggressive quantization, SNN model accuracy can be maintained while still profiting from both cheaper arithmetic and quantization-induced sparsity, thereby outperforming alternative model compression techniques. Additional fine-grained control can be achieved by further trading-off energy and accuracy by employing hybrid pruning and quantization schemes to deliver multiple models that occupy the accuracy-efficiency frontier.

\bibliographystyle{IEEEtran}
\bibliography{ref} 

\end{document}